# The Geography of Information Diffusion in Online Discourse on Europe and Migration

**Elisa Leonardelli, Sara Tonelli**

Fondazione Bruno Kessler, Trento, Italy
eleonardelli@fbk.eu, satonelli@fbk.eu

## Abstract

The online diffusion of information related to Europe and migration has been little investigated from an external point of view. However, this is a very relevant topic, especially if users have had no direct contact with Europe and its perception depends solely on information retrieved online. In this work we analyse the information circulating online about Europe and migration after retrieving a large amount of data from social media (Twitter), to gain new insights into topics, magnitude, and dynamics of their diffusion. We combine retweets and hashtags network analysis with geolocation of users, linking thus data to geography and allowing analysis from an "outside Europe" perspective, with a special focus on Africa. We also introduce a novel approach based on cross-lingual quotes, i.e. when content in a language is commented and retweeted in another language, assuming these interactions are a proxy for connections between very distant communities.

Results show how the majority of online discussions occurs at a national level, especially when discussing migration. Language (English) is pivotal for information to become transnational and reach far. Transnational information flow is strongly unbalanced, with content mainly produced in Europe and amplified outside. Conversely Europe-based accounts tend to be self-referential when they discuss migration-related topics. Football is the most exported topic from Europe worldwide. Moreover, important nodes in the communities discussing migration-related topics include accounts of official institutions and international agencies, together with journalists, news, commentators and activists.

## 1 Introduction

In recent years, the widespread use of smartphones and social media has revolutionized information access. Concurrently, all the information available online plays a critical role in forming opinions, perceptions, and impressions, especially about distant realities, for which direct experience is not available. Investigating information circulating online about a given subject can thus shed a new light into how it is perceived and why.

*Information diffusion* refers to the spreading of an idea or a piece of news through communication channels (Rogers 2010). With the rising popularity of online social networks, started around 2007, they have attracted a lot of attention from researchers that aimed to study the process of information diffusion through these channels. Indeed, online social networks pose new challenges when studying the process of large-scale information diffusion, since they have been found to deviate from the known characteristics of human social networks (Kwak et al. 2010) and to follow different rules. For example, if on the one hand they overcome geographic constraints, they are still bounded by a number of factors such as language, the online social context in which each user is immersed, the activity of users (e.g. choosing to follow someone or retweet some news etc.), the recommendation algorithms behind every specific social network, etc.

Aim of the current study is to investigate the online diffusion of information in relation to specific subjects, i.e. *Europe and migration to it*. Crucially, we want to study them using large amounts of data retrieved from Twitter[1] and adopt an "external point of view", i.e. from outside Europe. Indeed, Europe can be identified as a geographical location, and as such, it is possible to disentangle the position of users with respect to it. But Europe is also a concept, especially in the absence of a direct contact, and as such its perception can depend on the related information available online. Although several studies have addressed these topics, they were always addressed with a Europe-centric vision (e.g. Khatua and Nejdl 2021; Siapera et al. 2018), while to the best of our knowledge this is the first study to use a different perspective. The goal of this study is thus two-fold: we want to investigate online information diffusion about *Europe* outside its geographical boundaries, to understand how its external image is shaped. Furthermore, we aim at studying it in connection to *migration*.

As a matter of fact, according to the World Migration Report 2022,[2] Europe is still the major destination for international migrants (87 million migrants, 30.9% of the international migrant population). Previous studies, mainly based on direct interviews to migrants, showed how one of the main goals of choosing EU is the desire to improve the standard of living, mainly connected to find safety, freedom and education (McMahon and Sigona 2018). However, researchers observed also how often the decision to migrate was based more on vague hopes and ideas, rather than on actual knowledge about life in Europe (Carling and Sagmo 2015).

---

[1]We still use the name 'Twitter' throughout the paper because the data was collected in 2021 before the platform became 'X'.

[2]https://publications.iom.int/books/world-migration-report-2022

Information coming from online sources can exert its influence on a pre-migration phase, by shaping the imagination and perceptions of how migration and life in other countries might be (Dekker, Engbersen, and Faber 2016; Thulin and Vilhelmson 2014). Only if one can imagine an improvement in their condition in a new country, it can grow the ambitions to make this happen (De Haas 2011, 2014, 2021). Indeed, social media have been found to have a deep impact on the migration process, in the during- and post-migration phase (Dekker and Engbersen 2014) and an increasing amount of research has been devoted to understanding the relation between migration and digital technologies (Leurs and Smets 2018; Sîrbu et al. 2021). Studying the global spreading of information about Europe and migration to it is therefore crucial to understand to what extent online sources have an impact on aspirations and ambitions. We are interested in assessing the magnitude of online discussions around Europe and migration, how these are broadcast, and which are the factors that contribute to their diffusion. We also analyse how information flows from country to country, beyond national borders, across continents, shaping the perception of what is outside one's own country and possibly influencing ambitions and aspirations of future migrants. To this end, an essential part of this study is the geolocation of users, that links data from social media to geographical information so to analyze information flows inside and outside Europe.

This article is structured as follows. In Section 2 we describe how our Twitter dataset was created and we provide basic statistics about it, including the geolocation distribution of the accounts. In Section 3 we study how information related to Europe and migration is circulating on Twitter from a geographical perspective, looking at the location of producers (i.e. users who post original content) and amplifiers (i.e. those who retweet or comment). This analysis is further detailed in Section 4, where we focus on online conversations between European and African countries. In Section 5, we present a novel approach: we consider cross-lingual quotes i.e., when content in a language is commented and retweeted in another language. We assume these interactions to be particularly meaningful as they can connect users that belong to very distant communities and do not speak the same language. Moreover, in Section 6 we investigate the role of football, as the analysis of the previous sections highlighted a prominent role of this topic in our dataset. Finally, we summarize our findings and discuss future research directions in Section 7.

## 2 Dataset

### 2.1 Selection of the social media to monitor

In 2021, when we started our analysis, Twitter had more than 320 million active users globally and enabled concise conversations online with a light moderation policy. A relevant number of scientific works has traditionally focused on this platform, since it was one of the few social media that allowed access to public data via free API, before updating its terms of use in 2023 when it was renamed X. Each tweet could be retrieved with a set of more than 40 fields containing metadata pertaining the tweet itself (date, type, virality, etc.) and information about the account such as screen name and number of followers. Moreover, the fact that Twitter is globally used, with messages written wordlwide in multiple languages, allows us to capture information flows beyond national borders.

### 2.2 Data collection

A list of relevant query terms to retrieve tweets was manually defined by a panel of international domain experts in the partner consortium of the PERCEPTIONS European project,[3] who selected keywords and popular hashtags related to migration (e.g. *migrants*, *#UNHCR*, *Lampedusa*, *#Fortresseurope*, *#refugeecrisis*, etc.) as well as expressions specifically referring to migration such as *Barca ou barsakh* ('Barcelona or death'), a reference to migration used in Senegal. As we are interested also in more generic mainstream media conversations related to Europe, without specific reference to migration, a second more generic type of keywords was included for monitoring purposes, e.g. *Europe* and *#Eu*. A list of around 70 keywords, initially created in English, was then translated by the same pool of experts into five other languages: Arabic, French, German, Italian, and Spanish. These 420 multilingual keywords list is used as query terms to retrieve all tweets containing at least one of such keywords. Tweets were collected through Twitter stream API for a period of 3 months between 1st April 2021 and 30th June 2021. The collected tweets are about 32.5 million and contain 7.8 millions of unique accounts. The 60 most-frequent keywords are listed in Table 1 in the Appendix.

**Topic distribution** Figure 1 (left) shows the distribution of tweets in the collected dataset divided by topic. Topics are defined by manually grouping query-terms into Europe-related (i.e., trigger terms were such as *#Eu* or *Europe* or *Europa* etc.) or migration-related. A third case is when keywords from both topics are present. These tweets, for the sake of simplicity, have been considered Europe-related in the following analyses.

**Language distribution** Figure 1 (right) shows the language distribution in the collected dataset. To identify the language of a tweet we rely on the information provided by Twitter as metadata associated with each message. Although the keywords we use as query terms are only in 6 languages, no constraints were set on the language of the tweet to be downloaded. Indeed Figure 1 shows the presence of a variety of languages.

---

[3] https://www.perceptions.eu/

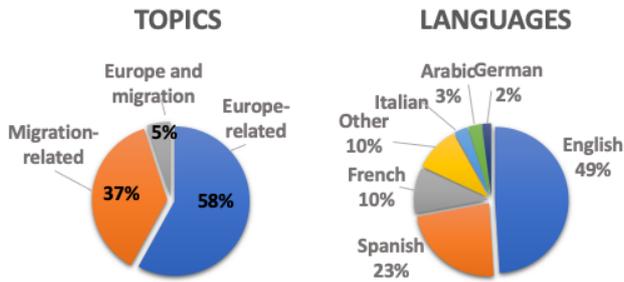

Figure 1: Overview of the distribution of topics and languages in the collected data

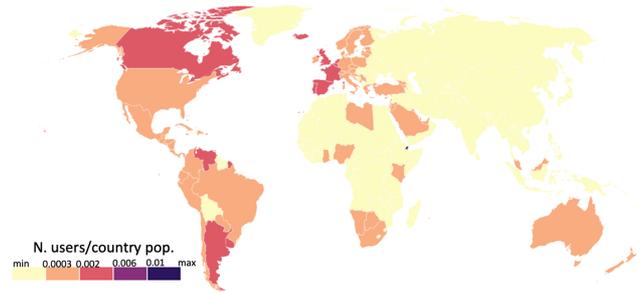

Figure 2: Distribution of localised users in our data, normalized by the population of each country

### 2.3 User localization procedure

Retrieving information about the geographical location of users in the dataset is crucial for the topic under scrutiny. Inferring it though, is one of the most difficult challenges when working with Twitter data (Zohar 2021), especially because since mid-2019, Twitter removed the option to share a tweet with geolocated coordinates, which was usually the preferred method used to estimate location (Kruspe et al. 2021). We therefore had to rely on the location information in the users' profile description, when not empty, to estimate users' location in our dataset. However, the content of the location field shows a high variability, since users can fill it with a free text string specifying their nationality, country of origin, country of residence, or any other type of information reflecting their sense of belonging.

In the collected data, for 60% of total users (4.7 million) the 'location' field in the profile description is not empty. For this subset of users, we first low-case the text and convert each emoji, if any, in the corresponding textual description, similarly to what was done in Leonardelli, Menini, and Tonelli (2020). This last step is useful to locate users that put the emoji flag of their country as location. We then run a matching algorithm between location strings and a list of countries and cities downloaded from the website of the United Nations Statistics Division. In particular, we first run a match with the countries list, which leads to the identification of 37% of all locations. We then match the remaining unidentified locations' strings with the list of cities with more than 400,000 inhabitants, localizing an additional 22% of the entries. Overall, we estimate the provenance at a country level for 38% of all the users in the dataset. Although we are aware that this is only a portion of the users, we consider this sample large enough (3 million users) to provide useful additional insights and be considered a representative sample of the users' geographical distribution in our dataset.

**Evaluation of the localization procedure** To evaluate the quality of the localization procedure, a subset of 370 non empty locations fields has been randomly extracted from the data and manually annotated with a country label. By comparing the manually annotated data with the estimated geolocations, we observe that more than 86% of the assigned locations were correct, around 12% were missed identifications, i.e. no country was identified while on the contrary there was one specified in the location field, and only 1.6% were wrong assignments. As an additional analysis, we annotate the 370 location fields by assigning one of three options: "no location", when the text in the field does not specify any location; "exact location", when the text describes only an existing location; "embedded" location, when the location is expressed through flags emojis, coordinates, slang, adjectives, or multiple locations are specified. We observe that around 20% of the location fields do not specify any location, and in this case we consider our algorithm correct if no country is assigned. Around 71% of the location fields contain an exact location, while around 9% contain some information on the location but in an embedded form. The most common source of error in our geolocalization algorithm is due to ambiguous cases, such as the Colombian city "Cartagena of Indias", that our algorithm assigned to India. Instead, our algorithm performs particularly well when identifying embedded locations.

### 2.4 Geographical distribution of users

Figure 2 shows the distribution of users' locations that we were able to identify in our dataset, normalized by the population size of each country. The majority of users is European (39%) as expected, given the fact that we used query terms related to this continent, such as "Europe". The most represented European countries are Great Britain, Spain and France (30%, 20%, 18% of the European users respectively). About 9% of users are located in Africa, with the most represented African countries in our dataset being Nigeria, South Africa, Kenya, Egypt and Ghana (34%, 15%, 9%, 8%, 7% respectively). Moreover, about 21% of the users are from North America, with United States being the most represented (66%) followed by Mexico (13%) and Canada (12%). About 16% are from South America (the most represented countries being Brazil and Argentina).

It is interesting to note that the distribution map in Fig. 2 recalls quite closely the Twitter users distribution map of Hawelka et al. (2014). This suggests that the provenance of users in our dataset strongly depends on the distribution of Twitter users in the world. Indeed, the incidence of users from Nigeria, South Africa and Ghana is not due to a large share of migrants to Europe compared to other African countries, but rather to the more widespread use of Twitter.

## 3 *Producers*, *amplifiers* and the geography of information flow

Retweets are a fast way to spread information and a key mechanism for information diffusion on Twitter (Jansen et al. 2009; Suh et al. 2010), which is usually associated with an agreement on the content of the original tweet and with endorsement (Boyd, Golder, and Lotan 2010; Lee, Hwalbin, and Okhyun 2015). A retweet always involves two users: one that we can define the *producer*, i.e. the person who writes an original content and shares it on Twitter, and the *amplifier*, i.e. the person retweeting the original message without adding any content but giving it visibility. Through this chain of actions, the message of the original *producer* can reach a wider audience and gain greater visibility. Concerning migration phenomena and Europe-related discourse, analysing where *producers* and *amplifiers* are located provides a better understanding of where this kind of information originates and where these messages gain more impact and visibility.

In order to analyse this aspect, in this section we focus on the part of the dataset consisting of retweets in which both users have been localised. The retweets for which both the location of producer and amplifier is known are about 5.1M. Although this represents only 12% of the original dataset, it can still be considered a relevant sample, from which meaningful insights can be obtained.

### 3.1 Information flow across countries

Despite the total absence of geographical boundaries in the digital world, in our dataset the majority of retweets (63%) occurs between two accounts that are localized in the same country (*intranational*), while only the remaining 37% of retweets are between users in different countries (*transnational*). This is probably due to the fact that a shared language within national borders facilitates online exchanges, but also confirms early studies on Twitter, which showed that users following each other are more likely to be geographically close, especially if they have 1,000 followers or less (Kwak et al. 2010).

To further refine this analysis, we split the dataset into two parts: one is the part of tweets collected through keywords related to migration, while the other pertains to tweets containing keywords about Europe. Interestingly, Figure 3 shows how the percentage of intranational/transnational retweets varies across the two subsets: migration-related discussion has higher percentage of intranational retweets, suggesting the discussion is more at internal level, possibly political, compared to Europe-related discussion. Since one of the goals of this study is to understand how information generates and spreads across borders, we further investigate transnational exchanges. Transnational exchanges between users within the same continent are only 13% of the retweets, suggesting a limited effect of geographical proximity of users. On the contrary, the language of the tweet exerts an important influence on how much a message will be retweeted outside the country. Indeed, 67% of the transnational retweets are in English, followed by Spanish (18%) and French (6%).

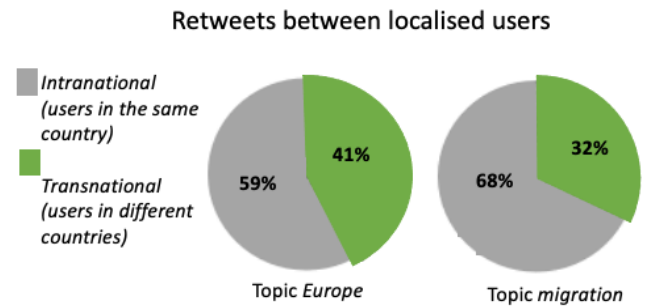

Figure 3: Share of producers and amplifiers located in the same country (Intranational) and in different countries (Transnational) for tweets with different topics: migration is a topic discussed more at a national level compared to Europe

In Figure 4 the transnational flow of messages is further analysed focusing on their direction and size. Again, we compare transnational discourse on "Europe" (1.1 million retweets) and on "migration" (600,000 retweets). Countries have been grouped into macro areas,[4] whose colors are displayed in the map at the top, and transnational retweets have been analyzed considering *i)* the macro-areas where producers and amplifiers were located (top vs. bottom half of the graphs) and *ii)* the amount and direction of retweets between *producers* and *amplifiers* localized in different areas. At the bottom of each graph, a bar with the users' distribution in the dataset (regardless of whether they are producers or amplifiers) is presented.

The left graph shows that most of the transnational conversations about Europe originates from producers located within this continent. If we consider the users' distribution, we notice that European users are overrepresented in the producers' bar, while continents such as Africa and Asia are underrepresented. Conversely, the latter are amplifiers, given that the second largest group of amplifiers are accounts located in Africa. This shows how much interest Europe-related discourse receives in the African continent. As regards migration-related tweets, we observe that producers are more balanced across continents compared to Europe-related tweets. However, the picture is still very unbalanced, with areas such as Europe and North America being very dominant in the production and being almost self-referential at amplifiers level. Between these two continents there are several exchanges related to migration: European users tend to amplify tweets issued by North American users on Central America, Iraq, Palestine and Brexit. On the other hand, North American amplifiers tend to retweet European producers when they discuss migration policies and initiatives affecting asylum seekers as well as Brexit and events in Palestine. Other areas, for example Africa, represent only a small portion in terms of producers when discussing migration-related topics, while they are exposed and receptive to messages coming from abroad.

---

[4]The macro areas selection is taken from https://unstats.un.org/unsd/methodology/m49/

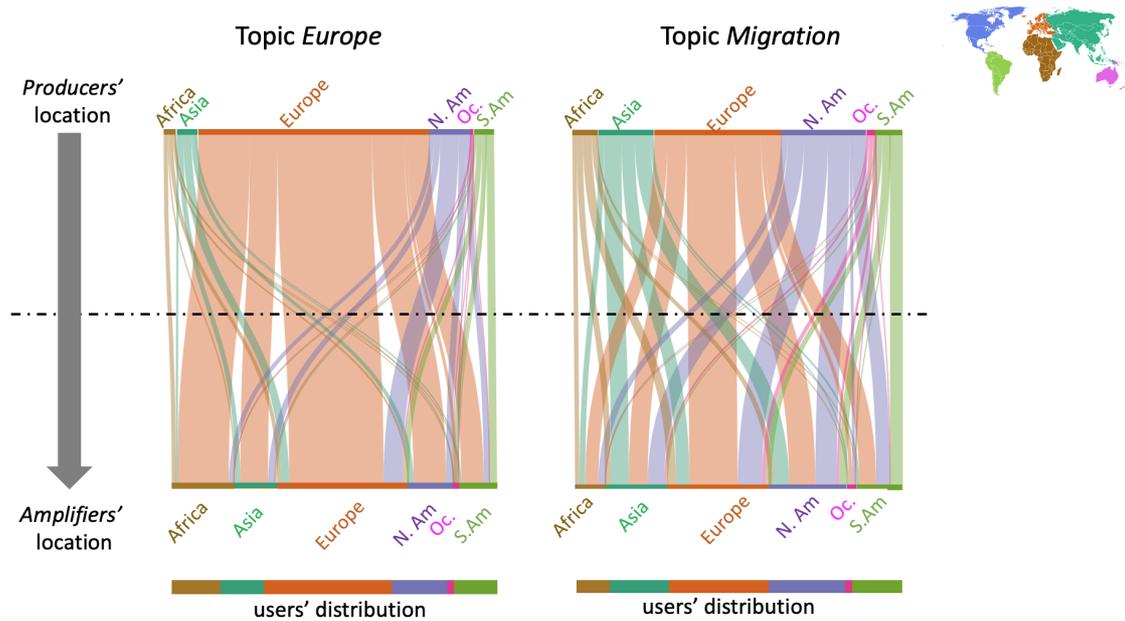

Figure 4: Transnational retweets related to "Europe" and "migration". The users' distribution is computed by merging all producers and amplifiers and counting only unique users.

## 4 Exchanges between Africa and Europe

To shed further light into the information exchange between Europe and Africa, we focus our analysis on retweets where the two users involved (producer and amplifier) are located one in Africa and one in Europe. This selection leads to about 220,000 retweets, generated by about 35,000 original tweets. Most of the selected data consists of tweets where the producers are located in Europe and retweeted by users in Africa (87% of retweets, 71% of the original tweets). These unbalanced numbers are already representative of this dataset, where a larger amount of content is created in Europe and spread in Africa rather than the other way round.

To further analyse the dynamics and structure of such exchanges, we build a retweet network, consisting of 1,927 nodes (*producers*) and 4,795 edges (*amplifiers*), where the weight of the nodes reflects the number of times an account has been retweeted (and thus its influence). To identify online communities in our dataset, we apply the Louvain algorithm (Blondel et al. 2008). We obtain 26 communities (modularity value of 0.845). We display the resulting network in Fig. 5. There, the colors of the communities match those reported in Fig. 6, where we provide an overview of the first 10 communities, that account for 65% of the total nodes. To identify the topic of discussion of each community, we first isolated the tweets generated by each community. We then performed an additional graph-analysis on the hashtags contained in the tweets for each community, using the hashtags as nodes and their co-occurences as edges. The most influential hashtags of the hashtag-graph obtained for each community, together with the most 10 influential ac-counts, were used to identify the main topic(s) in each community. We also capture the languages and the location of the users in each community (countries are shown in order of relevance within a community and only countries that account for more than 5% are shown). Finally, by observing these characteristics and the most relevant nodes (which are not reported in this work for privacy reasons), we manually assign a description to each community.

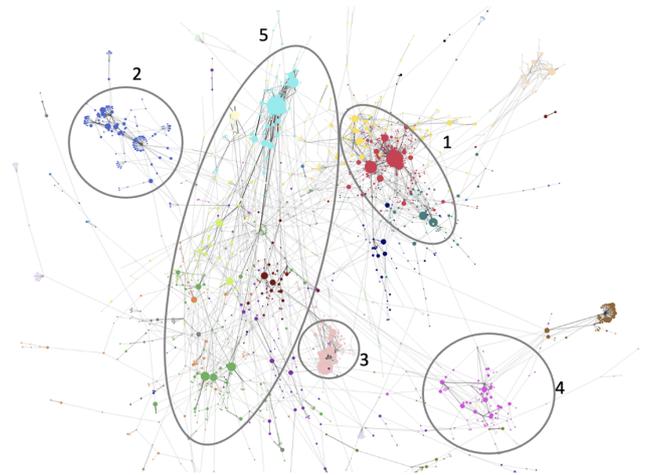

Figure 5: Retweet network between users in Europe and Africa with colorized communities.

This analysis confirms that Europe has a central role in the production of tweets, while Africa-based users mostly receive information and amplify messages. Indeed, none of the communities shown are composed of producers located only in Africa: producers are always either in Europe or in Europe and Africa. Furthermore, each amplifier network includes users located in Africa. The few communities where most producers are located in Africa discuss about the Palestinian situation, the Tigray/Ethiopia situation, and Western Sahara situation. These seem to be the topics with high visibility that stem from Africa and are spread in Europe. The most numerous communities with producers located mainly in Europe and with high visibility in Africa are those related to football, far apart from the other communities and close to each other (yellow and red clusters top-right, circled as number 1 in Figure 5). In this network, football represents a very popular topic of discussion. Indeed, two of the most numerous communities discuss topics around football and account for 30% of the nodes considered. Moreover, the ten most retweeted messages of the entire subset selected here are about football, and have been produced by football clubs, players, football-related news and football journalists. Specifically, they are about the Chelsea team winning the Europa League (which occurred at the end of May 2021, during our data collection period). Chelsea is a team known to include players of diverse nationalities and sees among their players several African ones. Indeed, the most retweeted messages often include pictures or mentions of such players. Besides football, the biggest cluster of communities (cluster 5) consists of a mix of agencies for human rights and their representatives, activists, journalists, news and commentators and Europe-related institutions and representatives. Indeed, information coming from governments and NGOs is increasingly exchanged through social media (Dekker, Engbersen, and Faber 2016). If we look at the main hashtags, we observe that these communities discuss topics that are closely related to each other such as human rights, climate change, covid-19, West Africa Competitiveness Programme (WACOMP), Universal Periodic Review (UPR) etc. The main language of discussion is English. If we focus on the most influential accounts, i.e. those that get the most retweets, we observe that in this community they include mostly accounts of real users, and not of institutions, for instance of *Ursula von der Leyen* or *Josep Borrell* for the EU-community, or UNHCR representatives in the human rights community. This suggests that online users are more likely to engage in a conversation or give visibility to real people, even if they have professional accounts, rather than to well-known and respected institutions. A French speaking community is also captured by the network (blue community, circle 2). There, tweets with migration-related hashtags are produced mainly in France and retweeted, besides France, in African countries that have historically tight relationships with France and where many can speak French (Morocco, Senegal, Togo, etc.). Interestingly, in all the communities, the most influential users consists of a mix of a few mass media sources and numerous *evangelists*, as defined by Cha et al. (2012). This term is used to refer to influential personalities such as leaders, politicians, celebrities, and journalists, who are able to reach audiences that are far away from the core of the network and introduce them to new topics.

| Community | Cluster | Network description | Main languages | Central Hashtags | Location of 'producers' -> 'amplifiers' | |
|---|---|---|---|---|---|---|
| 🟥 | 1 | Football-related | En 97% | #UCLFinal, #ChelseaChampions, #UCL | Europe -> Africa (99%) | |
| | | | | | (Gbr 79%, Ita 15%, ..) | (Nga 51%, Gha 21%, Ken 8%, Zaf 7%, ..) |
| 🟩 | 5 | Accounts related to EU official organisations and representatives | En 75%, Fr 20% | #EU, #WACOMP, #Europe, #COVID19 | Europe -> Africa (78%) | |
| | | | | | (Bel 42%, Fra 13%, Gbr 11%, ..) | (Ken 13%, Nga 11%, Zaf 6%, Gha 6%, ..) |
| 🟨 | 1 | Football-related | En 97% | #UEL, #mufc, #UELfinal, #AFC | Europe -> Africa (99%) | |
| | | | | | (Gbr 81%, Che 12%, ..) | (Nga 52%, Gha 18%, Ken 12%, Zaf 6%, ..) |
| 🟦 | 2 | French network: French TV and news, politicians, commentators | Fr 98% | #migrants, #Immigration, #France | Europe -> Africa (83%) | |
| | | | | | (Fra 80%, Tun 8%, ..) | (Fra 14%, Mar 12%, Sen 12%, Tgo 9%, Tun 8%, Dza 7%, Civ 7%, Cmr 5%, Cog 5%, ..) |
| 🟧 | 5 | Agencies for human rights observance and promotion, commentators, journalists, activists | En 92% | #humanrights, #UPR38, #refugees, #Syria, #Gaza, #Palestine | Africa -> Europe (75%) | |
| | | | | | (Pse 70%, Gbr 16%, ..) | (Gbr 41%, Fra 10%, Zaf 4%, ..) |
| 🟪 | 5 | Agencies for human rights observance and promotion an its representatives, journalists, news | En 66%, Fr 21%, Es 10% | #EU, #WACOMP, #Europe, #COVID19, #UE, #EuropeDay, #TeamEurope, #Africa, #AUEU | Europe -> Africa (65%) | |
| | | | | | (Fra 13%, Esp 13%, Bel 12%, Che 10%, Tun 9%, Gbr 9%, Ken 6%, Rwa 6%, ..) | (Gbr 9%, Uga 9%, Esp 9%, Civ 7%, Zaf 5%, Nga 5%, ..) |
| 🟦 | 5 | Agencies for human rights observance and promotion an its representatives, activists | En 97% | #Migration, #Africa, #ClimateChang, #Europe, #EhoA, #migration, #MigrationGovernance, #Libya, #Sudan | Europe -> Africa (69%) | |
| | | | | | (Che 52%, Deu 10%, Lby 6%, Ken 5%, ..) | (Nga 8%, Zaf 7%, Ken 7%, Gbr 7%, Che 6%, Eth 6%, ..) |
| 🟪 | 4 | Moroccan-based network composed by activists, news, commentators | En 44%, Fr 43%, Ar 11% | #WesternSahara, #Morocco, #polisario, #HumanRights, #UN, #SouthAlgeria, #MoroccanSaharawi, #Europe | Africa -> Europe (75%) | |
| | | | | | (Mar 72%, Fra 8%, Ita 6%, ..) | (Fra 29%, Mar 17%, Deu 16%, Nor 14%, Bel 5%, ..) |
| 🟩 | 5 | Agencies for human rights promotion and their representatives, journalists | En 83% | #refugee, #migrants, #COVID19, #EU, #HumanRights, #migranti, #UNHCR | Europe -> Africa (58%) | |
| | | | | | (Che 34%, Gbr 10%, Lyb 10%, Bel 5%, ..) | (Gbr 12%, Ken 11%, Nga 8%, Ita 7%, Che 5%, ..) |
| 🟪 | 3 | Tigray situation. Activists, politicians, commentators, journalists | En 97% | #Eu, #US, #Ethiopia, #EritreaPrevails #TPLF, #Tigray | Africa -> Europe (67%) | |
| | | | | | (Eri 48%, Gbr 18%, Eth 15%, Nld 6%, ..) | (Gbr 22%, Swe 14%, Eth 13%, Deu 8%, Nld 8%, Nga 7%, Nor 5%, ..) |

Figure 6: Characterization of the ten most relevant communities. Note that, in the location field, between the two possible directions (*producers* in Europe and *amplifiers* Africa or vice versa) only the main direction is shown, while the other is the remaining percentage.

## 5 Cross-lingual outreach

Beside retweets, where the content of an online message is re-posted by an *amplifier* to give it more visibility and endorse its content, there is another form of re-posting, the quote-retweet. Introduced by Twitter in 2015, it represents a compromise between the intention to share content and to engage in a conversation. Indeed, a quote is an edited retweet, where the user reshares the post but adds a personal comment. A quote allows users to express their opinion in the context of the original tweet, not necessarily an agreeing one (Garimella, Weber, and De Choudhury 2016). Edited retweets could consist of as high as 30% of the total retweets (Mustafaraj and Metaxas 2011).

In the analysis presented in this section, we focus on *cross-lingual quotes*, that is messages that have been posted in a language and have been retweeted with a comment in another language (see example in Fig. 7). One interesting way of modeling social networks derives from the theory of Granovetter (1973), that characterizes them in terms of strong and weak ties, where strong ties refer to relations with close friends or relatives, while weak ties represent links with distant acquaintances or unknown people. One key feature of online social networks is that they expanded the number of possible weak ties (Grabowicz et al. 2012), and thus can form bridges and link individuals to other social circles that usually are estranged to them. Our assumption is that cross-lingual quotes can overcome language barriers and connect circles of users that can be very far. They represent thus very significant hubs and only the most powerful and relevant discussions will be able to find resonance through these hubs. Hence, cross-lingual quotes can suggest which are the most viral, engaging or controversial messages online. We consider it as a proxy to identify cultural influence between distant communities.

### 5.1 Cross-lingual quote analysis

For this analysis, we select from the complete dataset only cross-lingual quotes. About 350,000 tweets were identified as cross-lingual quotes, commenting about 168,000 original tweets. A first interesting observation is that 43% of the original tweets then quoted are produced by a verified user. Conversely, only the 0.2% of the cross-lingual quotes are done by verified users. This shows that verified users tend to be highly visible also internationally, so that the content they produce is often made accessible to users speaking other languages. 53% of the original tweets are in English, confirming the importance of using this language when we want to reach a large audience, followed by Spanish (12%), Portuguese (10%), French (8.9%) and Catalan (2.6%). The cross-lingual quotes map the original tweets into a large number of languages (n=61). Tweets in English are quoted into 60 languages (the most relevant are Spanish, French, Hindu, German, Italian, Dutch, Turkish, Arabic). Tweets in Spanish are quoted into 43 languages (the most frequent ones being English, Catalan, French, Italian, Indian, Portuguese, Estonian) and tweets in French are quoted in 47 languages (the most frequent ones being English, Spanish, Italian, Hindi).

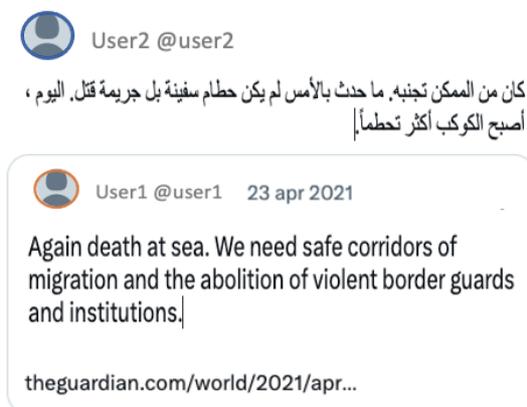

Figure 7: A (fictitious) example of cross-lingual quote, showing a tweet posted in English and quoted in Arabic

### 5.2 Cross-lingual quote network

Using the dataset of cross-lingual quotes, we build a network where nodes represent Twitter accounts and edges correspond to cross-lingual quote relation. The resulting network consists of 6,372 nodes and 11,834 edges. We identify influential communities (59 communities and a modularity value of 0.823) via Louvain algorithm (see Figure 8). The first 10 communities (52% of the total nodes) are detailed in Figure 9.

Three of the largest communities in the network discuss football-related topics, using hashtags concerning popular leagues and teams. In Fig. 8, they appear to be grouped together (upper right circle number 6), being rather separated from the rest of the network, and together consist of 17% of the total nodes. Extending the results of the previous analysis, this shows that discourse on European *football* starting in Europe has the potential to engage users in discussions beyond the borders of a specific team or nation at European level. Indeed, amplifiers are located basically in all continents.

Another topic generating discussions able to involve different countries is *music*, in particular the Eurovision song contest. Indeed, this event seems to be able to connect different countries within Europe and is popular also in Asia. This confirms that, similar to football, singers and musicians can contribute to create an image of Europe with an impact beyond European borders.

One of the largest communities in the cross-lingual quotes network is the one related to EU institutions and offices, to which some of the most influential accounts belong (circle 1). This depends largely on the keywords that were used to retrieve the tweets, since these accounts are mainly devoted to telling news and give updates on EU activities. As in the previous analysis, the personal accounts from institutional representatives are the most popular. Other topics that characterize the international communities discussing migration and EU are rescues in the Mediterranean Sea and Covid-19. For these communities, most of the producers and amplifiers are in Europe, suggesting that we are observing a discussion within European countries in the first place. Another community (blue, circle 3) revolves around populistic accounts that discuss Europe and migration. It is interesting to note how the main producers are located in Europe but also in North America (specifically USA), suggesting that this discussion concerns countries that are usually destinations for migrants. The community discussing Palestinian situation instead is mainly addressed by Africa-based users and, through quoting, reaches almost all continents. Overall, quoting networks tend to include a lot of media outlets and news channels, which have a very high number of followers. This confirms the findings of past studies, showing the key role played by mass media accounts in directly covering a large fraction of the Twitter audience, even if posting a relatively low number of tweets (Cha et al. 2012).

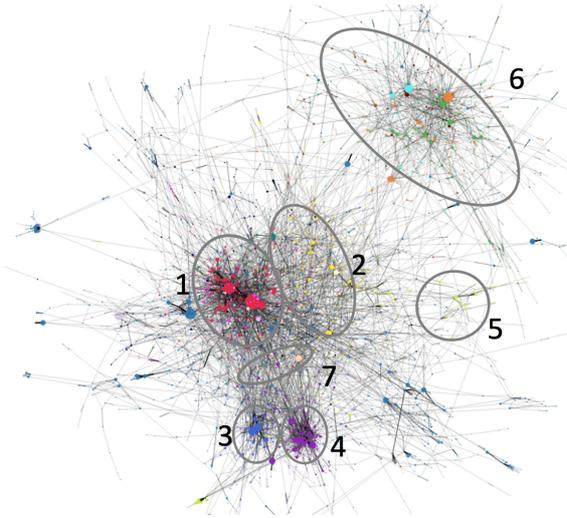

Figure 8: Cross-lingual quote network, with colors representing communities as assigned by the Louvain algorithm.

## 6 The role of football in the dataset

Since we observed in the previous analyses that football plays a relevant role in exchanges between Europe and Africa (Section 4) and in the cross-lingual quote network (Section 5), we further investigate the role of football in our dataset.

We first extract a random pool of around 2,000 tweets belonging to all the communities that were labelled as pertaining to football in the previous analysis and manually check them to compile a list of keywords related to football that were not part of the initial list of query terms. Using this approach, we obtain a list of 31 terms,[5] which we use to extract a subset of tweets from the original dataset containing only football-related content. By controlling for the presence of one (or more) football-related keyword, we isolate about 10% of the entire data collection described in Section 2 (4.3m of tweets), which is thus pertaining to the topic of football. 57% of the football-related tweets contain the term *league*, 28% the term *champion* and 17% the term *Chelsea*. By looking at which keywords originally triggered the download, we observe that 91% of the football-related dataset was originally downloaded because of the presence of the query term "Europe/Europa" (or its Arabic translation), used mostly to refer to "Europa League".

Figure 10 shows the time-series of the retrieved data related to football as well as the geographical distribution of users tweeting about this topic. There are clear peaks in relation to salient events, for instance the week of the UEFA Europa League finals accounts alone for around 29% of the tweets we isolated. This suggests that collecting our data while this big European event took place has surely impacted the content of the dataset, in particular the high volume of football-related tweets. Interestingly, the distribution of users in the football related dataset (Figure **??**) shows that online discourse on this topic reached far beyond the European borders: European football is a topic able to galvanize users all over the world and possibly influence the perception of Europe.

## 7 Discussion and Conclusions

In this work we analyze how discourse around Europe and migration is shaped on Twitter, with a focus on the "outside Europe" point of view. The goal is to study information diffusion and mechanisms that contribute to forming perceptions and aspirations about Europe and migration. The analysis of online processes is addressed by collecting a keyword-based dataset of 32 million tweets over a period of three months around two main subjects: Europe and migration. After geo-localising users in our dataset, we divide them into two groups, distinguishing between users that are content producers and users that are content amplifiers (i.e. that retweeted or quoted the producers' content). This chain of actions is at the basis of information flows on Twitter and shows how visibility to content is granted, underlying the importance of producers' point of views and narratives. In our first step described in Section 3, the geographical location of content producers and content amplifiers are put in connection, to infer the direction of information flow. Results show that, in our dataset, 63% of the total retweets occur at national level between producers and amplifiers located in the same country. These findings are in line with Kulshrestha et al. (2012), which estimate similar percentages in a study on information flows on Twitter. In our anal-ysis, we further show how this share changes depending on the topic: when discussing migration, interactions tend to be more at a national level than when discussing the subject "Europe". This is rather surprising because migration is a transnational topic almost by definition, but it has also a strong national connotation being tightly connected to local policies. Focusing on the retweets that cross national borders, we observe that several factors impact transnational exchanges, most importantly the language of the tweet (by far the majority of transnational tweets are in English) but also the Twitter penetration rate in the country of the amplifiers. Geographical proximity of countries, instead, has a marginal role in transnational retweeting. Moreover, when the subject is Europe, the main information "exporters" are users located in Europe itself. Interestingly, despite the fact that Africa-based accounts represent only 9% of localized users in our dataset, the second most numerous group of transnational amplifiers of Europe-related tweets is the African group, highlighting their interest in European events. Conversely, transnational exchanges about migration are more homogeneously distributed at global level because users in almost every country produce content on this subject. Furthermore, in Section 4 we isolate the data from users that are located either in Europe or Africa and dissect the online relationship between these two continents. We observe that the most relevant topic discussed by Europe-based users but gaining high visibility in African countries through retweets is football. This is in line with the fact that narratives related to successful football players with African origins seem to be quite popular in African countries, presenting a positive example of migrants, as found in field interviews (Esson 2015). Moreover, we find connections between countries that have linguistic and historical cultural bonds such as between France and Morocco, Senegal, Togo, etc. Finally, in Section 5 we present a methodologically novel analysis that investigates tweets that we define cross-lingual quotes, e.g. retweets with a comment in another language. We consider this type of interaction as a bridge between groups of users in the online network that are very far and thus a proxy for the most impactful and visible messages. This analysis can be seen as more generic with respect to the previous one since no limits are set on the user locations included in the considered dataset. Also in this analysis, English is by far the most influential source language, being the most quoted language, followed by Spanish and French, while the quoting languages are more diverse. Moreover, we observe that the majority of cross-lingual quotes involve verified accounts as producers. Also here, football-related tweets seem to be very popular and reach basically all continents. Interestingly, also Eurovision, i.e. the European music festival, seems to be a very popular topic able to connect users that are far apart. It has a particularly high visibility in discussions across European countries but also in Asia. This analysis suggests that football, Eurovision and messages from EU institutions are three main topics characterising the European public sphere (Hänska and Bauchowitz 2019), and playing a role in shaping the European identity conveyed outside this continent. What emerges from all the analyses is an unbalanced picture, with a strong predominance of producers located in first-world countries. For example, we observe that African countries are mostly receivers and amplify content from other continents. Our findings, however, could be influenced by other factors, for instance the online behaviour within language-specific communities on Twitter. Hong, Convertino, and Chi (2011) show, for instance, that users from different countries tend to use differently (i.e. more or less frequently) hashtags, user mentions and retweets. These cross-cultural differences may play a role in shaping the quoting behaviour in the different languages that we analyse.

---

[5]league, liverpool, chelsea, champions, ronaldo, juventus, psg, uefa, campiones, campioni, fcbarcelona, arsenal, seriea, futbol, calcio, football, villareal, worldcup, realmadrid, antonioconte, campeonas, mourinho, equipedefrance, bleus, gueye, manchestercity, بطولة، مانشستر، يوفنتوس، الدوري تشيلي

| Community | Cluster | Network main hubs description | Central hashtags | Locations of 'producer' | -> | Locations of 'amplifiers' | Languages of 'producer' | -> | Languages of 'amplifiers' |
|---|---|---|---|---|---|---|---|---|---|
| ■ | 1 | Related to EU official organizations | #EU, #Covid19, #EUCO | Europe 94 % | -> | Europe 85% | En 80%, Fr 6%, Es 5% | -> | Es 25%, Fr 12%, En 11%, De 9% |
| ■ | 6 | Football | #UCL, #UEL, #EURO2020 | Europe 92% | -> | Europe 44%, Africa 24%, Asia 16%, N.America 9% | En 72%, Fr 14%, Es 7% | -> | Es 21%, Fr 20%, In 13%, En 13% |
| ■ | 2 | International outlets | #EU, #HumanRights, #Europe, #US | Europe 84 % | -> | Europe 80% | Es 30%, En 30%, Ca 29% | -> | Es 37%, Ca 21%, En 15%, Fr 6% |
| ■ | 3 | Populists | #Eu, #migrants, #Italy, #Merkel | Europe 82%, N.America 9% | -> | Europe 69%, N.America 10%, Asia 9% | En 65%, Fr 12%, It 9% | -> | Es 37%, Ca 21%, En 15%, Fr 6% |
| ■ | 6 | Football | #UCL, #MUFC, #UEL, #EU, #SempreMilan | Europe 94% | -> | Europe 49%, Asia 19%, Africa 15%, S.America 9% | En 50%, Es 33%, Fr 6% | -> | En 23%, Es 22%, Fr 15%, In 12% |
| ■ | 4 | NGOs | #migranti, #Lybia, #EU, #Lampedusa | Europe 84% | -> | Europe 79% | En 65%, It 13%, Es 7% | -> | De 24%, Es21%, En 15%, It 11% |
| ■ | 6 | Football | #EURO2020, #Nazionale, #Azzurri, #ITA, #VivoAzzurro | Europe 86% | -> | Europe 41%, S.Amer 23%, N.Amer 13% | En 42%, Fr 37%, pt 9% | -> | Es 35%, En16%, In 11%, Fr 10% |
| ■ | 1 | Human rights activists | #EU, #HumanRights, #Menschenrechte, #Belarus, #StandWithBelarus, #Russia | Europe 73%, N. America 17% | -> | Europe 83% | En 82%, De 5%, Fr 4% | -> | De 17%, Es 14%, Fr 14%, En 13% |
| ■ | 5 | Pro-palestinian Activists | #HumanRights, #FreePalestine #IsraeliCrimes | Africa 45%, Europe 28%, Asia 15% | -> | Europe 38%, Asia 30%, Africa 18% , N. America 11% | Ar 54%, En 37%, Es 5% | -> | En 50%, Fr 13%, Es 9%, Ar 7% |
| ■ | 7 | Eurovision contest | #Eurovision, #OpenUp, #Eurovision2021 | Europe 88% | -> | Europe 75%, Asia 8% | En 81%, Es 8%, Fr 5% | -> | Es 36%, Fr 11%, It 10%, En 10% |

Figure 9: Ten most relevant communities in the cross-lingual quote network. To note that here, not all user locations are known, thus the percentage of users under "Locations of" refers only to the subset of localized users

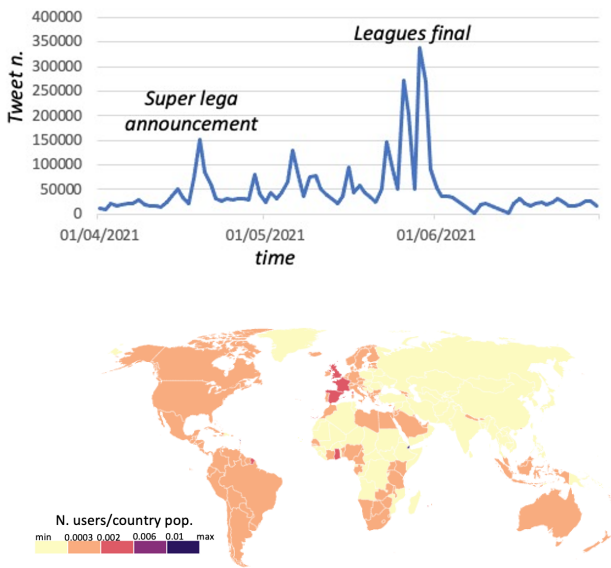

Figure 10: Timeline of football-related tweets and location of the users tweeting football-related content

The role of football emerged as one of the principal narratives contributing to the idea of Europe (within and outside it). Football clubs, football players, football news and football journalist are very influential nodes that convey messages resonating all around the globe and play a primary role in transmitting football-related migration narratives and success stories. Moreover, besides football, in both networks the most influential nodes are represented by official accounts of governments, institutions and agencies, especially official accounts of their representatives. For example, the personal account of Ursula Von Der Leyen was more pivotal in the network than the official accounts of EU institutions, together with journalists, politicians, commentators, activists, and news outlets, which are also influential hubs. In general, influential and highly visible communities seem to be based on a combination of different account types. It is finally interesting to note that overtly "anti-migrant" narratives, as well as "pro-migrant" ones (for example, the one containing the hashtag *#refugeesnotwelcome*, but also *#refugeeswelcome*) seem to be marginal. A populistic community emerges in the cross-lingual quote network but seems to capture a discussion between Europe and USA. As a side observation, European users are not very interested in information stemming from African users, limiting this information to few topics such as Palestinian territories or the Western Sahara situation.

**Broader perspective, ethics and competing interests**
Our work can be used to understand how Europe's perception is shaped online from outside the continent, and consequently which topics and events influence aspirations of potential migrants to Europe. Indeed, the relevance and high coverage of events such the Europa League and Eurovision song contest contribute to creating the image of Europe perceived also beyond its borders. The extensive visibility of events in Europe showcasing success stories might motivate migrants to seek opportunities in the continent, as they see the potential for a better quality of life. Our work can also be used to understand what kinds of communication strategies could be potentially successful in case of campaigns targeting countries outside Europe. For instance, this could involve having influential figures from the football industry to promote specific messages. On the contrary, more realistic portrayals of migration have less resonance than success stories.

Data collection has been carried out following a data-minimisation principle, retaining only the information strictly necessary to perform our analyses, as required by the Ethics board of the PERCEPTIONS project. The study has been carried out with a low granularity, aggregating users by country or by language. Similarly, the data released[6] with the current paper contains only the information necessary to replicate our study, without the possibility to retrieve the original data. Each tweet is represented as a (randomly assigned) ID, the tweet language, the user nation, the contained hashtags, the user ID (randomly assigned) and the retweets/quotes ID (randomly assigned). The textual content has been removed together with the real user names. The authors do not have any competing interests with this study.

**Limitations** A limitation of our work regards the selected social media. Indeed, Twitter users' distribution is skewed towards younger and richer individuals (Mislove et al. 2011), which might not be representative of the sample of population involved in migration events. However, it has been shown that migrants are largely young individuals and that there is a positive correlation between higher usage of Internet and rural-urban migration (Vilhelmson and Thulin 2013) and Internet access and migration in Africa (Grubanov-Boskovic et al. 2021). Moreover, Twitter as a social media has already been used for migration-related studies, such as to estimate international migration flows (Aswad and Menezes 2018; Hausmann, Hinz, and Yildirim 2018; Mazzoli et al. 2020; Zagheni et al. 2014), infer cross-border movements (Blanford et al. 2015), and analyse the integration of immigrants (Lamanna et al. 2018).

Another limitation concerns the fact that the 70 query terms were translated only in 6 languages, which are predominantly European languages. This could introduce biases in the captured data and in the represented population. However, the selected languages are among the most widely used on Twitter, with English alone representing more than 50% of tweets (Hong, Convertino, and Chi 2011). Moreover, keywords such as *#EU* and *Europe* are internationally used, and are often found in tweets in diverse languages, which would all be included in our dataset since we do not discard any language by default. The limited impact of this language-related bias is confirmed also by the similarity between our map of users distribution and the map of Twitter users in Hawelka et al. (2014), as well as the language rank by frequency, similar to the the language distribution in Hong, Convertino, and Chi (2011).

---
[6]https://github.com/dhfbk/ICWSM24-Geography-of-information-diffusion-online-Europe-Migration

A final limitation regards the fact that the data collected are limited to a specific, relatively short, three-month period. Although the period is long enough to allow meaningful observations, events like Chelsea winning the Europa League can skew the observed topics and massively influence online conversations taking place during this time period.

## Appendix

| Freq | Query term | Freq | Query term |
|---|---|---|---|
| 7.5M | europe | 98,052 | traficante |
| 6.8M | europa | 91,676 | asile |
| 1.1M | migrant | 89,835 | europe, رايس |
| 882,887 | asylum | 81,967 | lesbo |
| 767,104 | migration | 81,544 | regularization |
| 715,851 | #eu | 79,334 | legalizacion |
| 613,841 | #humanrights | 76,647 | trata de personas |
| 591,063 | #eu, europe | 74,510 | اللجوء |
| 469,573 | expulsion | 72,444 | #refugee, #refugees |
| 395,351 | human trafficking | 71,419 | |
| 380,760 | europa, europe | 70,665 | mediterraneo |
| 330,663 | أوروبا | 66,738 | migrant, Migration |
| 310,309 | refugee camp | 66,288 | europa, IOM |
| 259,648 | #eu, europa | 65,662 | غرباء |
| 247,088 | migracion | 65,333 | asyl, asylum, aigrant |
| 246,437 | asilo | 64,871 | regularizacion |
| 236,224 | NCR, UNHCR | 64,307 | europe, migrant |
| 208,485 | deportation | 62,776 | boza |
| 196,272 | #derechoshumanos | 58,281 | asylum, migration |
| 174,237 | amnistia | 53,445 | border control |
| 162,328 | #familiesBelong-together | 53,186 | asyl |
| 147,256 | #ue | 50,808 | europa,0IM |
| 131,212 | OIM | 50,419 | #eu,europa,europe |
| 123,992 | رايس | 50,401 | europe, migration |
| 119,462 | moria | 50,253 | asile, europa |
| 119,031 | IOM | 49,749 | lesbo, lesbos |
| 113,230 | nostalgia | 49,707 | #migrant, lfmigrants |
| 105,097 | traficante | 49,604 | lampedusa |
| 104,452 | #immigration | 49,155 | passeur |
| 99,117 | #ue, europa | 47,804 | أوروبا إبعاد |

Table 1: The 60 most frequent query terms present in our dataset, together with their number of occurrences.

## Acknowledgments

This research has been supported by the European Union's Horizon 2020 program projects PERCEPTIONS (GA 833870) and AI4TRUST (GA 101070190). We would like to thank Veronica Orsanigo for her help in evaluating the accuracy of the geolocation process.

## References

Aswad, F. M. S.; and Menezes, R. 2018. Refugee and Immigration: Twitter as a Proxy for Reality. In *Proceedings of FLAIRS Conference*, 253–258.

Blanford, J. I.; Huang, Z.; Savelyev, A.; and MacEachren, A. M. 2015. Geo-located tweets. Enhancing mobility maps and capturing cross-border movement. *PloS one*, 10(6): e0129202. Publisher: Public Library of Science San Francisco, CA USA.

Blondel, V. D.; Guillaume, J.-L.; Lambiotte, R.; and Lefebvre, E. 2008. Fast unfolding of communities in large networks. *Journal of Statistical Mechanics: Theory and Experiment*, 2008(10). _eprint: 0803.0476.

Boyd, D.; Golder, S.; and Lotan, G. 2010. Tweet, Tweet, Retweet: Conversational Aspects of Retweeting on Twitter. In *Proceedings of the 2010 43rd Hawaii International Conference on System Sciences*, HICSS '10, 1–10. IEEE Computer Society. ISBN 978-0-7695-3869-3.

Carling, J.; and Sagmo, T. 2015. Rumour and migration. *Benevenutonging Face of Global Mobility*.

Cha, M.; Benevenuto, F.; Haddadi, H.; and Gummadi, K. P. 2012. The World of Connections and Information Flow in Twitter. *IEEE Transactions on Systems, Man, and Cybernetics - Part A: Systems and Humans*, 42: 991–998.

De Haas, H. 2011. The determinants of international migration. Publisher: International Migration Institute.

De Haas, H. 2014. Migration theory: Quo vadis? Publisher: International Migration Institute.

De Haas, H. 2021. A theory of migration: the aspirations-capabilities framework. *Comparative Migration Studies*, 9(1): 8.

Dekker, R.; and Engbersen, G. 2014. How social media transform migrant networks and facilitate migration. 14(4): 401–418. Publisher: Wiley Online Library.

Dekker, R.; Engbersen, G.; and Faber, M. 2016. The use of online media in migration networks. 22(6): 539–551. Publisher: Wiley Online Library.

Esson, J. 2015. Better off at home? Rethinking responses to trafficked West African footballers in Europe. *Journal of Ethnic and Migration Studies*, 41(3): 512–530. Publisher: Taylor & Francis.

Garimella, K.; Weber, I.; and De Choudhury, M. 2016. Quote RTs on Twitter: usage of the new feature for political discourse. In *Proceedings of the 8th ACM Conference on Web Science*, 200–204. ACM. ISBN 978-1-4503-4208-7.

Grabowicz, P. A.; Ramasco, J. J.; Moro, E.; Pujol, J. M.; and Eguiluz, V. M. 2012. Social features of online networks: The strength of intermediary ties in online social media. *PloS one*, 7(1). Publisher: Public Library of Science San Francisco, USA.

Granovetter, M. S. 1973. The strength of weak ties. *American journal of sociology*, 78(6): 1360–1380. Publisher: University of Chicago Press.

Grubanov-Boskovic, S.; Kalantaryan, S.; Migali, S.; and Scipioni, M. 2021. The impact of the Internet on migration

aspirations and intentions. *Migration Studies*, 9(4): 1807–1822. Publisher: Oxford University Press.

Hausmann, R.; Hinz, J.; and Yildirim, M. A. 2018. Measuring Venezuelan emigration with Twitter.

Hawelka, B.; Sitko, I.; Beinat, E.; Sobolevsky, S.; Kazakopoulos, P.; and Ratti, C. 2014. Geo-located Twitter as proxy for global mobility patterns. *Cartography and Geographic Information Science*, 41(3): 260–271.

Hong, L.; Convertino, G.; and Chi, E. 2011. Language matters in twitter: A large scale study. In *Proceedings of the international AAAI conference on web and social media*, volume 5, 518–521.

Hänska, M.; and Bauchowitz, S. 2019. Can social media facilitate a European public sphere? Transnational communication and the Europeanization of Twitter during the Eurozone crisis. *Social media+ society*, 5(3): 2056305119854686. Publisher: SAGE Publications Sage UK: London, England.

Jansen, B. J.; Zhang, M.; Sobel, K.; and Chowdury, A. 2009. Twitter power: Tweets as electronic word of mouth. 60(11): 2169–2188. Publisher: Wiley Online Library.

Khatua, A.; and Nejdl, W. 2021. Analyzing European Migrant-related Twitter Deliberations. In *Companion Proceedings of the Web Conference 2021*, 166–170.

Kruspe, A.; Häberle, M.; Hoffmann, E. J.; Rode-Hasinger, S.; Abdulahhad, K.; and Zhu, X. X. 2021. Changes in Twitter geolocations: Insights and suggestions for future usage. In *Proceedings of the Seventh Workshop on Noisy User-generated Text (W-NUT 2021)*, 212–221. Association for Computational Linguistics.

Kulshrestha, J.; Kooti, F.; Nikravesh, A.; and Gummadi, K. 2012. Geographic dissection of the Twitter network. In *Proceedings of the International AAAI Conference on Web and Social Media*, volume 6, 202–209. Issue: 1.

Kwak, H.; Lee, B.; Park, H.; and Moon, S. 2010. What is Twitter, a social network or a news media? In *Proceedings of the 19th international conference on World wide web*, 591–600.

Lamanna, F.; Lenormand, M.; Salas-Olmedo, M. H.; Romanillos, G.; Gonçalves, B.; and Ramasco, J. J. 2018. Immigrant community integration in world cities. *PloS one*, 13(3): e0191612. Publisher: Public Library of Science San Francisco, CA USA.

Lee, M.; Hwalbin, K.; and Okhyun, K. 2015. Why do people retweet a tweet? Altruistic, egoistic and reciprocity motivations for retweeting. 58(4): 189–201.

Leonardelli, E.; Menini, S.; and Tonelli, S. 2020. DH-FBK@ HaSpeeDe2: Italian Hate Speech Detection via Self-Training and Oversampling. In *Evalita 2020-Seventh Evaluation Campaign of Natural Language Processing and Speech Tools for Italian*, volume 2765.

Leurs, K.; and Smets, K. 2018. Five questions for digital migration studies: Learning from digital connectivity and forced migration in (to) Europe. 4(1): 2056305118764425. Publisher: SAGE Publications Sage UK: London, England.

Mazzoli, M.; Diechtiareff, B.; Tugores, A.; Wives, W.; Adler, N.; Colet, P.; and Ramasco, J. J. 2020. Migrant mobility flows characterized with digital data. *PloS one*, 15(3): e0230264. Publisher: Public Library of Science San Francisco, CA USA.

McMahon, S.; and Sigona, N. 2018. Navigating the Central Mediterranean in a time of 'crisis': Disentangling migration governance and migrant journeys. *Sociology*, 52(3): 497–514.

Mislove, A.; Lehmann, S.; Ahn, Y.-Y.; Onnela, J.-P.; and Rosenquist, J. 2011. Understanding the demographics of Twitter users. In *Proceedings of the International AAAI Conference on Web and Social Media*, volume 5, 554–557. Issue: 1.

Mustafaraj, E.; and Metaxas, P. T. 2011. What Edited Retweets Reveal about Online Political Discourse. In *Proceedings of the Twenty-Fifth AAAI Conference on Artificial Intelligence*, 6.

Rogers, E. M. 2010. *Diffusion of innovations*. Simon and Schuster.

Siapera, E.; Boudourides, M.; Lenis, S.; and Suiter, J. 2018. Refugees and network publics on Twitter: Networked framing, affect, and capture. *Social Media+ Society*, 4(1): 2056305118764437. Publisher: SAGE Publications Sage UK: London, England.

Suh, B.; Hong, L.; Pirolli, P.; and Chi, E. H. 2010. Want to be retweeted? large scale analytics on factors impacting retweet in twitter network. In *2010 IEEE second international conference on social computing*, 177–184. IEEE.

Sîrbu, A.; Andrienko, G.; Andrienko, N.; Boldrini, C.; Conti, M.; Giannotti, F.; Guidotti, R.; Bertoli, S.; Kim, J.; Muntean, C. I.; and others. 2021. Human migration: the Big Data perspective. *International Journal of Data Science and Analytics*, 11(4): 341–360. Publisher: Springer.

Thulin, E.; and Vilhelmson, B. 2014. Virtual practices and migration plans: A qualitative study of urban young adults. 20(5): 389–401. Publisher: Wiley Online Library.

Vilhelmson, B.; and Thulin, E. 2013. Does the Internet encourage people to move? Investigating Swedish young adults' internal migration experiences and plans. *Geoforum*, 47: 209–216. Publisher: Elsevier.

Zagheni, E.; Garimella, V. R. K.; Weber, I.; and State, B. 2014. Inferring International and Internal Migration Patterns from Twitter Data. In *Proceedings of the 23rd International Conference on World Wide Web*, WWW '14 Companion, 439–444. Association for Computing Machinery. ISBN 978-1-4503-2745-9. Event-place: Seoul, Korea.

Zohar, M. 2021. Geolocating tweets via spatial inspection of information inferred from tweet meta-fields. *International Journal of Applied Earth Observation and Geoinformation*, 105: 102593.